\pdfoutput=1

\documentclass[11pt]{article}

\usepackage{ACL2023}

\usepackage{times}
\usepackage{latexsym}
\usepackage{multirow}
\usepackage{multicol}
\usepackage{graphicx}
\usepackage{booktabs}
\usepackage{subfigure}

\usepackage{tikz}
\usepackage{tikz-qtree}
\usetikzlibrary{arrows,decorations.pathmorphing,backgrounds,positioning,fit,petri,shapes.misc, arrows.meta,shapes.geometric,decorations.markings,calc,shadows.blur,decorations.pathreplacing,quotes}
\definecolor{myblue}{RGB}{6, 82, 221}
\definecolor{myorange}{RGB}{211, 84, 0}
\definecolor{lowblue}{RGB}{102,178,255}
\definecolor{justblue}{RGB}{84, 160, 255}
\definecolor{mypurple}{RGB}{108, 92, 231}
\definecolor{mygray}{RGB}{158, 158, 158}
\definecolor{lowpurple}{RGB}{204,153,255}
\definecolor{lowwhite}{RGB}{255,255,255}
\definecolor{verylowpurple}{RGB}{255,102,102}
\definecolor{embcolor}{RGB}{255,255,255}
\definecolor{myred}{RGB}{235, 47, 6} 
\definecolor{mygreen}{RGB}{162, 217, 206} 
\definecolor{fontgrey}{RGB}{44, 62, 80}
\definecolor{lowpurple}{RGB}{210, 180, 222}
\definecolor{mypumpkin}{RGB}{229, 152, 102}
\definecolor{lowgreen}{RGB}{171, 235, 198}
\definecolor{lowgreen2}{RGB}{186, 220, 88}
\definecolor{lowred}{RGB}{245, 183, 177}
\definecolor{lowyellow}{RGB}{241, 196, 15}
\definecolor{mypink}{RGB}{255, 118, 117}
\definecolor{bluemartina}{RGB}{18, 203, 196}
\definecolor{puffin}{RGB}{250, 152, 58}
\definecolor{grass}{RGB}{0, 148, 50}
\definecolor{cnngray}{RGB}{116, 125, 140}
\usepackage{pgfplots}
\usepackage{pgfpages}

\usepackage[T1]{fontenc}

\usepackage[utf8]{inputenc}

\usepackage{microtype}
\usepackage{arydshln}

\usepackage{inconsolata}

\newcommand*{\changefonthint}{\fontfamily{qcr}\selectfont}
\newcommand{\hint}[1]{{\changefonthint{#1}}}

\newcommand{\sel}[1]{{\changefonthint{#1}}}
\newcommand{\dataset}[1]{\texttt{#1}}
\newcommand{\red}[1]{\textcolor{red}{#1}}

\newcommand{\blue}[1]{\textcolor{blue}{#1}}
\newcommand{\brown}[1]{\textcolor{brown}{#1}}

\usepackage{CJKutf8}

\title{Easy-to-Hard Learning for Information Extraction\thanks{\hspace{1mm} This work was supported by Alibaba Group through Alibaba Research Intern Program. It was also partially supported by a grant from the Research Grant Council of the Hong Kong Special Administrative Region, China (Project Code: 14200719). This work was done when Chang Gao was an intern at Alibaba DAMO Academy.}}

\author{Chang Gao\textsuperscript{\rm 1,2}, Wenxuan Zhang\textsuperscript{\rm 2\footnotemark[2]}, Wai Lam\textsuperscript{\rm 1}, Lidong Bing\textsuperscript{\rm 2} \\
\textsuperscript{\rm 1}The Chinese University of Hong Kong \\ \textsuperscript{\rm 2}DAMO Academy, Alibaba Group \\
\texttt{\{gaochang,wlam\}@se.cuhk.edu.hk}  \\
\texttt{\{saike.zwx,l.bing\}@alibaba-inc.com} } 

\begin{document}
\maketitle
\renewcommand{\thefootnote}{\fnsymbol{footnote}}
\footnotetext[2]{Wenxuan Zhang is the corresponding author.}
\renewcommand{\thefootnote}{\arabic{footnote}}
\begin{abstract}
Information extraction (IE) systems aim to automatically extract structured information, such as named entities, relations between entities, and events, from unstructured texts. While most existing work addresses a particular IE task, universally modeling various IE tasks with one model has achieved great success recently. Despite their success, they employ a one-stage learning strategy, i.e., directly learning to extract the target structure given the input text, which contradicts the human learning process. In this paper, we propose a unified easy-to-hard learning framework consisting of three stages, i.e., the easy stage, the hard stage, and the main stage, for IE by mimicking the human learning process. By breaking down the learning process into multiple stages, our framework facilitates the model to acquire general IE task knowledge and improve its generalization ability.  Extensive experiments across four IE tasks demonstrate the effectiveness of our framework. We achieve new state-of-the-art results on 13 out of 17 datasets. Our code is available at \url{https://github.com/DAMO-NLP-SG/IE-E2H}.
\end{abstract}

\section{Introduction}
Information extraction (IE) is a crucial task in natural language processing (NLP) that involves extracting structured knowledge from unstructured text data \cite{bing-etal-2013-wsdm,bing-etal-2015-improving}, enabling various applications such as information retrieval \cite{IR}, knowledge graph construction \cite{KG,wang-etal-2019-tackling}, and question answering \cite{QA}. 
Depending on what kind of information is to be extracted, IE consists of a wide range of tasks, including named entity recognition (NER) \cite{NERSurvey}, joint entity and relation extraction (RE) \cite{taille-etal-2020-lets,chia-etal-2022-relationprompt}, event extraction (EE) \cite{EESurvey}, and aspect-based sentiment analysis (ABSA) \cite{ABSASurvey}.

Traditionally, IE has been approached with specialized models that are designed to handle specific IE tasks. For example, NER is often formulated as a sequence labeling \cite{ma-hovy-2016-end,xu-etal-2021-better} or span-based classification \cite{wang-etal-2020-pyramid} problem.
The more complex RE or EE task is usually solved with pipeline approaches that split the original task into several sequential subtasks and design specific models for each subtask \cite{subburathinam-etal-2019-cross, yang-etal-2019-exploring,DBLP:conf/aaai/PengXBHLS20}. 
These models often require extensive task-specific knowledge to design dedicated model architectures and thus suffer from poor generalization. Recently, motivated by pre-trained generative models such as T5 \cite{T5} that handle multiple tasks with the unified text-to-text format, there has been a shift towards the use of unified models for IE as well, which can tackle all IE tasks with a single model structure. For example, TANL \cite{TANL} tackles various IE tasks with a text-to-text generative model by framing them as translation between augmented natural languages. UIE \cite{UIE} models heterogeneous IE structures into a uniform representation via a structural extraction language.

Despite the success of existing unified models on various IE tasks, they typically adopt a one-stage learning paradigm, i.e., directly learning to predict the target structure given the input text. In contrast, humans often learn to tackle a task in an easy-to-hard manner.
They learn basic concepts or skills before solving more complex problems and often tackle harder examples to gain a better understanding of the problem. Taking the RE task as an example, it aims to extract relational triplets, where each triplet consists of a head entity, a relation, and a tail entity. To tackle it, humans first learn some basic skills, such as identifying entities, recognizing relations, and associating entities and relations, before extracting complex relational triplets. This process facilitates humans to learn meaningful substructures and the dependencies among them. Moreover, in practical scenarios, humans usually encounter harder cases, i.e., long input context of multiple sentences containing more entities and relations. By solving hard cases, humans improve their understanding of the task and problem-solving skills. By comparison, models are only trained with the provided training data.
The gap between the model and human learning strategies hinders IE models from further development.

To bridge the gap, we propose an \textbf{easy-to-hard (E2H)} learning framework for IE tasks in this paper. E2H mimics the human learning procedure to learn each IE task in stages, i.e., the easy stage, the hard stage, and the main stage. The easy stage aims to help the model acquire basic skills of the task, and the hard stage aims to assist the model in handling broad-range variations of the task via training the model with diverse and harder data. Finally, the main stage focuses on the main task at hand for training. Thus an immediate question is how to prepare the data with different levels of difficulty for the easy and hard stages. It is labor-intensive and challenging to construct such data manually. In this work, we attempt only to leverage the existing data of the main task for constructing the data.

Specifically, for the easy stage, we observe that the target IE structure often has meaningful substructures. Therefore, we identify several basic skills for each task according to the substructures of its target structure. Returning to the RE example, the skills can be recognizing the entities, relations, and dependencies between them. 
We can automatically construct training data for learning these skills by modifying the input prompt and decomposing the target structure of the main task. For the hard stage, we combine two training instances of the main task to build a harder training instance by concatenating their input texts to form the new text and their targets to build the new target. The new instance contains more entities, relations, and complicated contexts, making it harder than the original instances. Through these two novel construction strategies, we can reduce much human effort to obtain the data for different stages.

To summarize, our contributions are three-fold: (1) We propose a unified easy-to-hard (E2H) learning framework for IE tasks by imitating the human learning process; (2) We develop two novel strategies to build the easy and hard stages of our framework without using any additional resources;
(3) We conduct comprehensive evaluations on 17 datasets across four IE tasks and achieve state-of-the-art results on 13 datasets. Notably, our E2H method consistently outperforms the one-stage learning counterpart by introducing two extra learning stages  with an average increase of 0.38, 2.96, 1.33, and 1.39 absolute points on the NER, RE, EE, and ABSA tasks, respectively.

\section{Task Definition}
\begin{figure*}[tb]
  \centering
  \includegraphics[width=\linewidth]{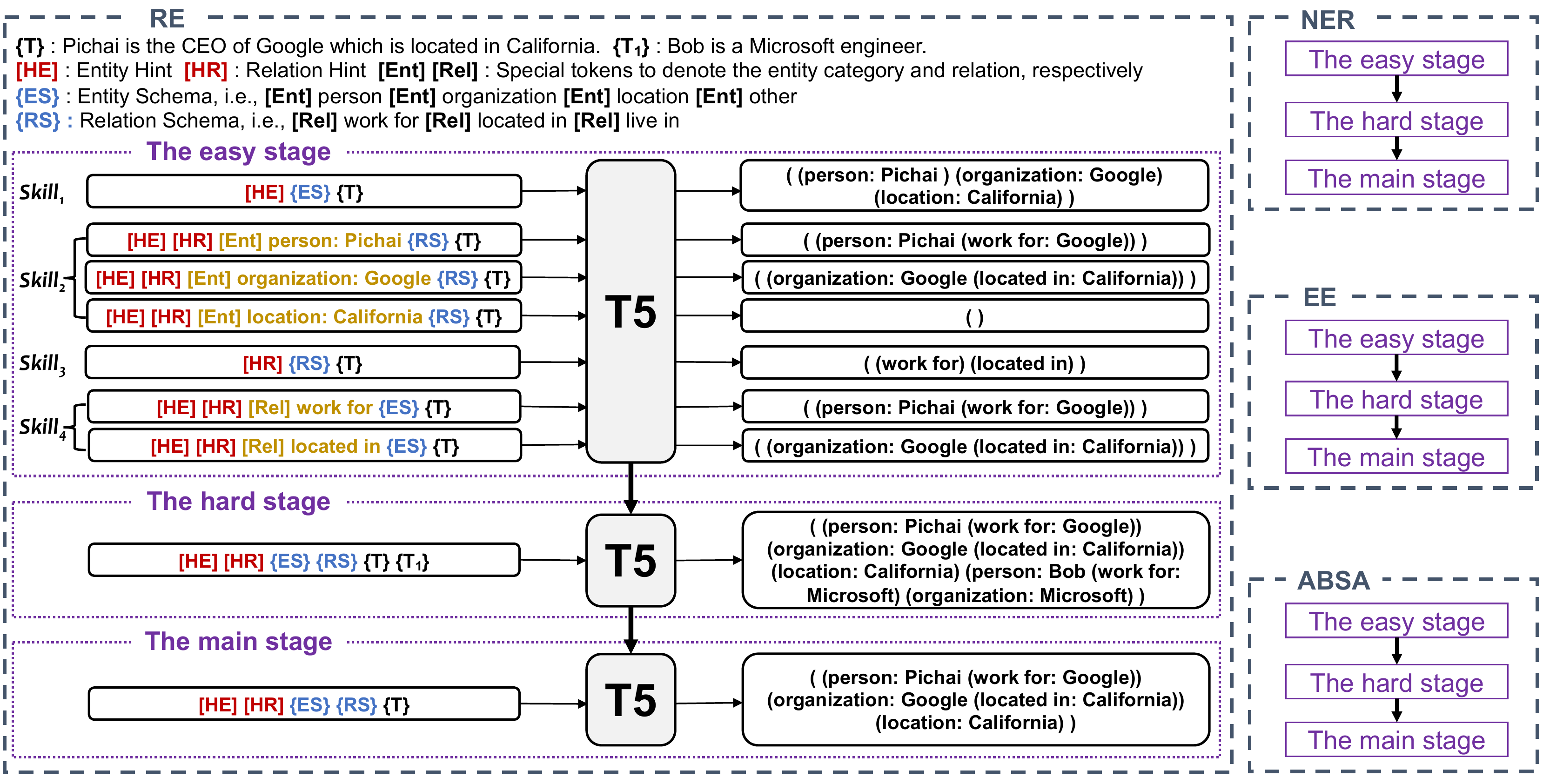} 
  \caption{Overview of E2H consisting of three stages, i.e., the easy stage, the hard stage, and the main stage. We highlight \hint{Hint} in \red{red}, \hint{Constraint} in \brown{brown}, and \hint{Schema} in \blue{blue}.}
  \label{fig:E2H}
\end{figure*}

This paper investigates four common IE tasks, i.e., NER, RE, EE, and ABSA. In this section, we provide formal definitions of these tasks. Detailed examples of these tasks are in Appendix \ref{sec:examples}.

\paragraph{Named Entity Recognition (NER)}
Given an input text $T$, the task is to identify and classify entities in $T$ into predefined categories, i.e., extract
$ \{(e_i, c_i)\}$, where $e_i$ is the $i$-th entity, which is a continuous text span in $T$, $c_i \in \mathcal{C}$ is its category, and $\mathcal{C}$ is the entity category set.

\paragraph{Relation Extraction (RE)}
Given an input text $T$, RE is to identify a set of (head entity, relation, tail entity) triplets, i.e., extract $\{ ((e^h_i, c^h_i), r_i, (e^t_i, c^t_i) )\}$, where the superscripts $h$ and $t$ denote the head and tail entities, $r_i \in \mathcal{R}$ is the $i$-th relation, and $\mathcal{R}$ is the relation set. 

\paragraph{Event Extraction (EE)}
Given an input text $T$, the task is to identify a set of events where each event consists of an event trigger and a set of corresponding arguments, i.e., extract $\{ \left((e^{tri}_i, c^{tri}_i), (e^{arg_1}_i, c^{arg_1}_i), \cdots, (e^{arg_m}_i, c^{arg_m}_i) \right)\}$, where $e^{tri}_i$ is the $i$-th trigger, which is a continuous text span in $T$, $c^{tri}_i \in \mathcal{C}_{event}$ is its category, $e^{arg_j}_i$ is the $j$-th argument of the $i$-th event, which is also a continuous text span in $T$, $c^{arg_j}_i \in \mathcal{C}_{event}$ is its category, and $\mathcal{C}_{event}$ consists of all event and argument categories.

\paragraph{Aspect-based Sentiment Analysis (ABSA)}
There are four essential elements in ABSA, namely aspect category $c$, aspect term $a$, opinion term $o$, and sentiment polarity $p$. We focus on the aspect sentiment triplet extraction (ASTE) task \cite{DBLP:conf/aaai/PengXBHLS20} and the aspect sentiment quad prediction (ASQP) task \cite{paraphrase} given their popularity. 
Given an input text $T$, the ASTE task is to identify a set of 
 $\{(a_i, o_i, p_i)\}$ triplets, and the ASQP task is to identify a set of $\{(c_i, a_i, o_i, p_i)\}$ quadruplets, where $c_i \in \mathcal{C}_{absa}$ is $i$-th aspect category, $a_i$ is $i$-th aspect term, $o_i$ is $i$-th opinion term, both $a_i$ and $o_i$ are continuous spans in $T$, $p_i\in \{\textrm{positive, negative, neutral}\}$ is $i$-th sentiment polarity, and $\mathcal{C}_{absa}$ is the aspect category set.

\section{Our E2H Framework}
Our proposed easy-to-hard (E2H) framework consists of three sequential stages: the easy stage, the hard stage, and the main stage. In this section, we first introduce our text-to-structure formulation for facilitating three-stage learning in a unified framework. Next, we will describe how to realize the easy and hard stages. Finally, we will discuss the main stage as well as the detailed training and inference process of our framework. 

\begin{table*}
\centering
\resizebox{\textwidth}{!}
{
\begin{tabular}{cl}
\toprule
\textbf{Task} & \textbf{Basic Skills} \\
\midrule
\multirow{2}{*}{NER} 
& \textit{Skill$_1$}: $T$ $\rightarrow$ a set of entity categories $\{c_i\}$ \\
& \textit{Skill$_2$}: $T$ and an entity category constraint $c$ $\rightarrow$ a set of entities of $c$ $\{(e_i, c)\}$ \\
 \midrule
 \multirow{4}{*}{RE} 
 & \textit{Skill$_1$}: $T$ $\rightarrow$ a set of entities $\{(e_i, c_i)\}$ \\
 & \textit{Skill$_2$}: $T$ and a head entity constraint $(e^h,c^h)$ $\rightarrow$ a set of relational triplets $\{((e^h, c^h), r_i, e^t_i)\}$ \\
 & \textit{Skill$_3$}: $T$ $\rightarrow$ a set of relations $\{r_i\}$ \\
 & \textit{Skill$_4$}: $T$ and a relation constraint\ $r$ $\rightarrow$ a set of relational triplets $\{((e^h_i, c^h_i), r, e^t_i)\}$ \\
 \midrule
 \multirow{2}{*}{EE} 
 & \textit{Skill$_1$}: $T$ $\rightarrow$ a set of event triggers $\{(e^{tri}_i, c^{tri}_i)\}$ \\
  & \textit{Skill$_2$}: $T$ and a trigger constraint $(e^{tri},c^{tri})$ $\rightarrow$ the event $\left((e^{tri}, c^{tri}), (e^{arg_1}, c^{arg_1}), \cdots, (e^{arg_m}, c^{arg_m}) \right)$  \\
 \midrule
\multirow{4}{*}{ASTE}
 & \textit{Skill$_1$}: $T$ $\rightarrow$ a set of aspect terms $\{a_i\}$ and a set of opinion terms $\{o_i\}$ \\
 & \textit{Skill$_2$}: $T$ and an aspect term constraint $a$ $\rightarrow$ a set of triplets $\{(a, o_i, p_i)\}$ \\
 & \textit{Skill$_3$}: $T$ $\rightarrow$ a set of sentiment polarities $\{p_i\}$ \\
& \textit{Skill$_4$}: $T$ and a sentiment polarity constraint $p$ $\rightarrow$ a set of triplets $\{(a_i, o_i, p)\}$ \\
\midrule
\multirow{4}{*}{ASQP} 
& \textit{Skill$_1$}: $T$ $\rightarrow$ a set of aspect categories $\{c_i\}$ \\
 & \textit{Skill$_2$}: $T$ $\rightarrow$ a set of (aspect category, aspect term) tuples $\{(c_i, a_i)\}$ \\
  & \textit{Skill$_3$}: $T$ $\rightarrow$ a set of (aspect category, opinion term) tuples $\{(c_i, o_i)\}$ \\
   & \textit{Skill$_4$}: $T$ $\rightarrow$ a set of  (aspect category, sentiment polarity) tuples $\{(c_i, p_i)\}$ \\
 \bottomrule
\end{tabular}
}
\caption{Basic skills for NER, RE, EE, ASTE, and ASQP. We omit \hint{Hint} and \hint{Schema} for simplicity. Detailed examples are in Appendix \ref{sec:examples}.}
\label{tab:skill}
\end{table*}

\subsection{Unified Text-to-Structure Formulation}

Similar to UIE \cite{UIE}, we formulate NER, RE, EE, and ABSA as text-to-structure generation problems, which allows us to use a single model to tackle multiple tasks. Given a text $T$ and its corresponding prompt $P$, we aim to generate the target IE structure $S$ with an encoder-decoder model $M:(P,T)\rightarrow S$. To facilitate the learning of different stages, we design the prompt $P$ containing three types of information: \hint{Hint}, \hint{Constraint}, and \hint{Schema}. \hint{Hint} guides the model on what elements should be extracted, \hint{Constraint} indicates specific constraints for the task, and \hint{Schema} provides necessary information such as the possible relation set for the extraction. With these three types of information, the prompt is able to connect the learning process in different stages. 

Taking the RE task as an example, as depicted in Figure \ref{fig:E2H}, \hint{Hint} consists of one or both of an entity hint and a relation hint. The entity hint, represented by the special token \hint{[HE]}, guides the model to extract entities, and the relation hint, represented by the special token \hint{[HR]}, guides the model to extract relations. The use of both hints guides the model to extract both entity and relation information, in the form of (head entity, relation, tail entity) triplets. \hint{Constraint} is a specific entity or relation, which limits the target structure to be related to that entity or relation. Lastly, \hint{Schema} contains pre-defined entity categories or relations or both of them, depending on the information that needs to be extracted. It provides essential information for identifying entities and relations in a text.

\subsection{The Easy Stage}
The goal of the easy stage is to enable the model to learn basic skills that will aid in tackling the main task. To achieve this, we identify several skills for each task and automatically construct the training data for them based on the data of the main task. 
Table \ref{tab:skill} presents the basic skills of NER, RE, EE, ASTE, and ASQP.  We design each skill to be a subtask of the main task according to its target structure. These skills are more fundamental and well-defined. Combining these skills gives the model a whole picture of how to tackle the main task. For example, the RE task has four skills. Skill$_1$ and Skill$_3$ help the model recognize substructures of the relational triplet, i.e., the entity and relation, respectively, and Skill$_2$ and Skill$_4$ help the model learn the dependencies between these substructures. 

To construct the training data for each skill, we modify the input and target of the main task's training data. Specifically, the input text is the same for the skills and the main task, but the prompt is different. As shown in Figure \ref{fig:E2H}, for the RE task, there is only \hint{[HE]} in the hint of Skill$_1$ as it only extracts entities and only \hint{[HR]} in the hint of Skill$_3$ as it only extracts relations. Both \hint{[HE]} and \hint{[HR]} are in the hints of Skill$_2$, Skill$_4$, and the main task because they extract (head entity, relation, tail entity) triplets. 
For Skill$_2$ and Skill$_4$, there is also a \hint{Constraint}, i.e., a head entity or relation, which requires their targets to be triplets related to a specific head entity or relation. The schema of the RE task consists of both entity categories and relations. For a specific skill of RE, the schema only contains entity categories or relations. The target of each skill is a part of the target of the RE task. For Skill$_1$ and Skill$_3$, which extract a substructure of the relational triplet, we use the substructure as the target. For Skill$_2$ and Skill$_4$, we use the corresponding subset of triplets of the RE task as the target.

\subsection{The Hard Stage}
The hard stage aims to construct training examples that are harder than the original training examples of the main task to train the model. Intuitively, the training instance is harder if the input text contains more structural elements and more complicated contexts. To this end, we combine two training instances of the original task to construct a harder instance. Formally, given two training instances $(P, T_1, S_1)$ and $(P, T_2, S_2)$, we can construct a harder training instance $(P, T_1 \circ T_2, S_1 \circ S_2 )$, where $P$ is the prompt, $T_i$ is the $i$-th text, $S_i$ is the $i$-th target structure, and $\circ$ denotes concatenation. An example is shown in the hard stage part of the RE task in Figure \ref{fig:E2H}. The model has to process and understand the combined information from both instances, making it more challenging for the model to correctly extract the target structure. 

Let $N$ denote the number of training examples of the original task. For each training example, we randomly sample $M$ training examples whose target structures are not empty to construct $M$ hard instances. This results in a total of $N*M$ hard instances. This approach allows us to easily construct a large amount of diverse hard training data.

\subsection{The Main Stage}

After training the model in the easy and hard stages, we train the model with the main task in this stage. 

\paragraph{Training} 

We adopt the pre-trained sequence-to-sequence model T5 \cite{T5} as the backbone of E2H. The model is trained with a maximum likelihood objective. Given the training example $(P,T,S)$, the loss function $L_\theta$ is defined as
\begin{equation}
L_{\theta}=-\sum_{i=1}^{n} \log P_{\theta}\left(S_{i} \mid S_{<i}, P, T \right)
\end{equation}
where $\theta$ is the model parameters, $P$ is the prompt, $T$ is the text, $S$ is the target structure, and $n$ is the length of $S$. 
We train the model in the easy, hard, and main stages sequentially. For the easy stage, we adopt the weights of pre-trained T5 to initialize the model. For the hard and main stages, we initialize the model with the weights of the model trained in the previous stage.

\paragraph{Inference}
Once the training process is complete, we use the model trained in the main stage to generate the target structure $S$ for any given tuple of the prompt and text $(P, T)$. Although our training process has three stages, the inference is a one-stage process. The computational load is the same as that of the one-stage learning counterpart.

\section{Experiments}

\subsection{Experimental Setup}
\paragraph{Datasets} 
We conduct experiments on 17 datasets across four IE tasks, i.e., NER, RE, EE, and ABSA. We evaluate the flat NER task with CoNLL03  \citep{tjongkimsang2003conll}, and the nested NER task with ACE04-Ent \citep{ace2004-annotation} and ACE05-Ent \citep{ace2005-annotation}. For RE, we experiment on CoNLL04  \citep{roth-yih-2004-linear}, ACE05-Rel \citep{ace2005-annotation}, and SciERC \citep{luan-etal-2018-multi}. Regarding to EE, we use ACE05E, ACE05E+ \citep{ace2005-annotation}, and CASIE \citep{Satyapanich_Ferraro_Finin_2020}. As for ABSA, we consider the ASTE and ASQP tasks. For ASTE, we adopt four popular datasets, including Rest14, Laptop14, Rest15, and Rest16 provided by \citet{xu-etal-2020-position}. 
For ASQP, we use R-ACOS and L-ACOS provided by \citet{ACOS}, and Rest15 and Rest16 provided by \citet{paraphrase}.
These ABSA datasets are derived from the datasets provided by the SemEval ABSA challenges \citep{pontiki-etal-2014-semeval, pontiki-etal-2015-semeval, pontiki-etal-2016-semeval}, except L-ACOS which is collected from the Amazon Laptop domain. Statistics of these datasets are provided in Appendix \ref{sec:datasets}. 

\paragraph{Evaluation} 
We use Micro-F1 as the primary evaluation metric. For each experimental result, we report the average performance on three random seeds. For NER, RE, EE, and ASTE, we follow \citet{UIE} to use Entity F1, Relation Strict F1, Event Trigger F1 and Argument F1, and Sentiment Triplet F1 as the evaluation metrics and map the generated string-level extraction results to offset-level for evaluation. For ASQP, we follow \citet{paraphrase} to use Sentiment Quad F1 to evaluate the model. A sentiment quad is correct if and only if the four elements are exactly the same as those in the gold sentiment quad.

\paragraph{Baselines} 
We divide our baselines into two categories: specialized models and unified models. Specialized models are designed for a particular IE task, while unified models are designed for general IE. For specialized models, we use state-of-the-art methods such as BARTNER \cite{BARTNER} and DeBias \cite{zhang-etal-2022-de} for NER, UniRE \cite{wang-etal-2021-unire} and PURE \cite{zhong-chen-2021-frustratingly} for RE, Text2Event \cite{lu-etal-2021-text2event} and DEGREE \cite{hsu-etal-2022-degree} for EE, and PARAPHRASE \cite{paraphrase} and Seq2Path \cite{mao-etal-2022-seq2path} for ABSA. For unified models, we use TANL \cite{TANL}, UIE \cite{UIE}, and LasUIE \cite{LasUIE} as baselines. To make a fair comparison with one-stage learning methods, we also build T5-base and T5-large baselines.  We set their inputs and outputs the same as those of E2H and only train them in the main stage. 

\paragraph{Implementation Details}
E2H has two model sizes: E2H-base and E2H-large, which are initialized with pre-trained T5-base and T5-large models \cite{T5}, respectively. Other details are reported in Appendix \ref{sec:implementation}.

\begin{table*}
\centering
\resizebox{\textwidth}{!}{
\setlength\tabcolsep{2pt}
\begin{tabular}{lcccccccccc}
\toprule
\multirow{2.5}{*}{\textbf{Models}} &
\multicolumn{4}{c}{NER} &  & \multicolumn{4}{c}{RE} & \\
\cmidrule{2-5}  \cmidrule{7-10}
 &  \dataset{CoNLL03 } & \dataset{ACE04-Ent} & \dataset{ACE05-Ent} & Avg & & \dataset{CoNLL04} & \dataset{ACE05-Rel} & \dataset{SciERC} & Avg \\
\midrule
\textit{Specialized Models}  \\
BARTNER \small{\cite{BARTNER}} & \bf{93.24} & 86.84 & 84.74 & 88.27 & &  - & - & -  & -\\
DeBias \small{\cite{zhang-etal-2022-de}} & 93.12 & 85.28 & 84.93 & 87.78 &  & - & - & -  & -\\
UniRE \small{\cite{wang-etal-2021-unire}} & - & - & - & - & &  - & 64.30 & \underline{36.90} & - \\
PURE \small{\cite{zhong-chen-2021-frustratingly}} & - & - & - & - & &  - & 64.80 & 36.80 & - \\
\midrule
\textit{Unified Models} \\
TANL \small{\cite{TANL}} & 91.70 & - & 84.90 & - &  & 71.40 & 63.70 & - & - \\
UIE$^*$ \small{\cite{UIE}} & 92.99 & \underline{86.89} & 85.78 & 88.55 &  & 75.00 & 66.06 & 36.53 & \underline{59.20} \\
LasUIE$^*$ \small{\cite{LasUIE}} & \underline{93.20} & 86.80 & \underline{86.00} & \bf{88.67} & & \underline{75.30} & \bf{66.40} & - & - \\
T5-base \small{\cite{T5}} & 91.72 & 85.60 & 84.16 & 87.16 & & 69.58 & 62.91 & 33.13 & 55.20\\
T5-large \small{\cite{T5}} & 92.05 & 86.78 & 85.76 & 88.20 & & 71.72 & 64.49 & 35.44 & 57.21\\
\hdashline
E2H-base  & 91.92  & 86.24 & 84.83 & 87.66 & & 72.23 & 65.44 & 35.06 & 57.58 \\
E2H-large & 92.43 & \bf{87.06} & \bf{86.25} & \underline{88.58} & & \bf{75.31} & \underline{66.21}  & \bf{39.00} & \bf{60.17} \\
\bottomrule
\end{tabular}
}
\caption{Experimental results on the NER and RE tasks. The best results are in bold and the second-best results are underlined. Models marked with $*$ conduct large-scale continued pre-training with external resources. Except for T5-base and T5-large, the results of baselines are taken from their original papers.}
\label{tab:NER}
\end{table*}

\begin{table*}
\centering
\resizebox{0.88\textwidth}{!}
{
\setlength\tabcolsep{2pt}
\begin{tabular}{lcccccccc}
\toprule
\multirow{2.5}{*}{\textbf{Models}} & \multicolumn{2}{c}{\dataset{ACE05-E}} & \multicolumn{2}{c}{\dataset{ACE05-E+}} & \multicolumn{2}{c}{\dataset{CASIE}} & \multicolumn{2}{c}{\dataset{Avg}}\\
\cmidrule{2-3}  \cmidrule{4-5}  \cmidrule{6-7} \cmidrule{8-9}
& Trig F1 & Argu F1 & Trig F1 & Argu F1 & Trig F1 & Argu F1 & Trig F1 & Argu F1  \\
\midrule
\textit{Specialized Models}  \\
Text2Event \small{\cite{lu-etal-2021-text2event}} & 71.90 & 53.80 & 71.80 & 54.40 & - & - & - & - \\
DEGREE \small{\cite{hsu-etal-2022-degree}} & \bf{73.30} & \bf{55.80} & 70.90 & \bf{56.30} & - & - & - & -  \\
\midrule
\textit{Unified Models} \\
TANL \small{\cite{TANL}} & 68.40 & 47.60 & - & - & - & - & - & -  \\
UIE$^*$ \small{\cite{UIE}} & - & - & \underline{73.36} & 54.79 & 69.33 & 61.30 & - & -  \\
T5-base \small{\cite{T5}} & 68.19 & 49.68 & 69.68 & 50.65 & 68.40 & 60.19 & 68.76 & 53.51  \\
T5-large \small{\cite{T5}} & 70.40 & 52.42 & 71.45 & 54.08 & 69.29 & 60.98 & \underline{70.38} & \underline{55.83} \\
\hdashline
E2H-base  & 70.12 & 50.98 & 69.99 & 52.85 & 68.45 & 60.40 & 69.52 & 54.74 \\
E2H-large & \underline{72.19} & \underline{53.85} & \bf{73.50} & \underline{55.67} & \bf{69.58} & \bf{61.96} & \bf{71.76} & \bf{57.16} \\
\bottomrule
\end{tabular}
}
\caption{Experimental results on the EE task. The best results are in bold and the second-best results are underlined.  Models marked with $*$ conduct large-scale continued pre-training with external resources. Except for T5-base and T5-large, the results of baselines are taken from their original papers.}
\label{tab:EE}
\end{table*}

\begin{table*}
\centering
\resizebox{\textwidth}{!}
{
\setlength\tabcolsep{2pt}
\begin{tabular}{lccccccccccc}
\toprule
\multirow{2.5}{*}{\textbf{Models}} & \multicolumn{5}{c}{ASTE} &  & \multicolumn{5}{c}{ASQP} \\
\cmidrule{2-6}  \cmidrule{8-12}
& \dataset{Rest14} & \dataset{Laptop14} & \dataset{Rest15} & \dataset{Rest16} & Avg & & \dataset{R-ACOS} & \dataset{L-ACOS} & \dataset{Rest15} & \dataset{Rest16} & Avg\\
\midrule
\textit{Specialized Models}  \\
PARAPHRASE \small{\cite{paraphrase}} & 72.03 & 61.13 & 62.56 & 71.70 & 66.86 & & - & - & 46.93 & 57.93 & - \\
Seq2Path \small{\cite{mao-etal-2022-seq2path}} & \underline{75.52} & 64.82 & 65.88 & 72.87 & 69.77 & & 58.41 & 42.97 & - & - & -\\
\midrule
\textit{Unified Models} \\
UIE$^*$ \small{\cite{UIE}} & 74.52 & 63.88 & 67.15 & \underline{75.07} & 70.16 & & - & - & - & - & - \\
T5-base \small{\cite{T5}} & 72.11 & 63.06 & 66.27 & 72.24 & 68.42 & & 59.26 & 43.12 & 48.24 & 58.92 & 52.39 \\
T5-large \small{\cite{T5}} & 73.48 & 63.62 & 67.08 & 74.85 & 69.76 & & \underline{61.24} & \underline{44.37} & \underline{51.76} & \underline{60.93} & \underline{54.58} \\
\hdashline
E2H-base  & 75.40 & \underline{65.78} & \underline{68.58} & 73.83 & \underline{70.90} & & 60.66 & 43.51 & 49.45 & 59.55 & 53.29 \\
E2H-large & \bf{75.92} & \bf{65.98} & \bf{68.80} & \bf{75.46} & \bf{71.54} &  & \bf{63.50} & \bf{44.51} & \bf{52.39} & \bf{61.86} & \bf{55.57}\\
\bottomrule
\end{tabular}
}
\caption{Experimental results on two ABSA tasks, including the ASTE task and the ASQP task.  The best results are in bold and the second-best results are underlined.  Models marked with $*$ conduct large-scale continued pre-training with external resources. Except for T5-base and T5-large, the results of baselines are taken from their original papers.}
\label{tab:ABSA}
\end{table*}

\subsection{Main Results}
 
We compare E2H with state-of-the-art specialized and unified models. Tables \ref{tab:NER}-\ref{tab:ABSA} report the experimental results on 17 datasets across four IE tasks. We have the following observations: (1) E2H is an effective framework for various IE tasks. E2H-large achieves new state-of-the-art results on 13 out of 17 datasets. (2) The proposed easy-to-hard three-stage learning method consistently outperforms the one-stage learning counterpart. E2H performs better than T5 on all the datasets for two model sizes, and E2H-large obtains an average improvement of 0.38, 2.96, 1.33, and 1.39 absolute points over T5-large  on the NER, RE, EE, and ABSA tasks, respectively. This demonstrates the strong generalization ability of our framework. (3) Without using any external resources, our method exhibits comparable or stronger performance than models with large-scale continued pre-training.  Compared with UIE \cite{UIE}, which is pre-trained with large-scale structured, unstructured, and parallel data, E2H-large achieves better performance on the RE, EE, and ASTE tasks and obtains comparable results on the NER task.  (4) Easy-to-hard learning brings more benefits to complex tasks than simple tasks. Specifically, compared with the improvement on the NER task, which only extracts entities, the improvements of E2H over T5 are more significant on the other three tasks, which extract tuples with multiple elements. This shows that our method can help the model effectively capture the structural dependency of complex structures.

\begin{figure}[tb]
\centering
\subfigure[NER results on ACE04-Ent]{
\centering
\begin{minipage}[t]{0.5\linewidth}
\centering
\begin{tikzpicture}
\pgfplotsset{width=4.8cm,height=4cm,compat=1.8}
\begin{axis}[
    xtick={1,2,3},
    xticklabels = {1\%, 5\%, 10\%},
    xticklabel style = {font=\fontsize{6}{1}\selectfont},
    yticklabel style = {font=\fontsize{6}{1}\selectfont},
    legend style={font=\fontsize{6}{1}\selectfont},
  xlabel={\tiny Data Ratio},
  enlargelimits=0.1,
  legend style={at={(0.7,0.05)},anchor=south,legend columns=1}, 
  every axis plot/.append style={thick},
  tick label style={/pgf/number format/fixed},
    every node near coord/.append style={font=\tiny}
]
 \addplot[lowyellow] [mark=triangle,mark size=2.7pt] coordinates {
(1, 56.4) (2, 75.16) (3, 79.51)
 };
 
\addplot[lowblue]  [mark=square]  coordinates {
(1, 43.8) (2, 68.06) (3, 75.96)
};

\legend{ {E2H-base}, {T5-base}}
\end{axis}
\end{tikzpicture}
\end{minipage}%
}%
\subfigure[RE results on ACE05-Rel]{
\centering
\begin{minipage}[t]{0.5\linewidth}
\centering
\begin{tikzpicture}
\pgfplotsset{width=4.8cm,height=4cm,compat=1.8}
\begin{axis}[
    xtick={1,2,3},
    xticklabels = {1\%, 5\%, 10\%},
    xticklabel style = {font=\fontsize{6}{1}\selectfont},
    yticklabel style = {font=\fontsize{6}{1}\selectfont},
    legend style={font=\fontsize{6}{1}\selectfont},
  xlabel={\tiny Data Ratio},
  enlargelimits=0.1,
  legend style={at={(0.7,0.05)},anchor=south,legend columns=1}, 
  every axis plot/.append style={thick},
  tick label style={/pgf/number format/fixed},
    every node near coord/.append style={font=\tiny}
]

 \addplot[lowyellow] [mark=triangle,mark size=2.7pt] coordinates {
(1, 23.9) (2, 46.66) (3, 54.6)	
 };
 
\addplot[lowblue]  [mark=square]  coordinates {
(1, 9.48) (2, 34.6) (3, 46.49)
};

\legend{ {E2H-base}, {T5-base}}
\end{axis}
\end{tikzpicture}
\end{minipage}%
}
\subfigure[EE results on ACE05-E]{
\centering
\begin{minipage}[t]{0.5\linewidth}
\centering
\begin{tikzpicture}
\pgfplotsset{width=4.8cm,height=4cm,compat=1.8}
\begin{axis}[
    xtick={1,2,3},
    xticklabels = {1\%, 5\%, 10\%},
    xticklabel style = {font=\fontsize{6}{1}\selectfont},
    yticklabel style = {font=\fontsize{6}{1}\selectfont},
    legend style={font=\fontsize{6}{1}\selectfont},
  xlabel={\tiny Data Ratio},
  enlargelimits=0.1,
  legend style={at={(0.7,0.05)},anchor=south,legend columns=1}, 
  every axis plot/.append style={thick},
  tick label style={/pgf/number format/fixed},
    every node near coord/.append style={font=\tiny}
]

 \addplot[lowyellow] [mark=triangle,mark size=2.7pt] coordinates {
 (1, 10.63) (2, 29.05) (3, 36.41)		
 };
 
\addplot[lowblue]  [mark=square]  coordinates {
(1, 5.4) (2, 22.64) (3, 31.28)
};

\legend{ {E2H-base}, {T5-base}}
\end{axis}
\end{tikzpicture}
\end{minipage}%
}%
\subfigure[ABSA results on Rest14]{
\centering
\begin{minipage}[t]{0.5\linewidth}
\centering
\begin{tikzpicture}
\pgfplotsset{width=4.8cm,height=4cm,compat=1.8}
\begin{axis}[
    xtick={1,2,3},
    xticklabels = {1\%, 5\%, 10\%},
    xticklabel style = {font=\fontsize{6}{1}\selectfont},
    yticklabel style = {font=\fontsize{6}{1}\selectfont},
    legend style={font=\fontsize{6}{1}\selectfont},
  xlabel={\tiny Data Ratio},
  enlargelimits=0.1,
  legend style={at={(0.7,0.05)},anchor=south,legend columns=1}, 
  every axis plot/.append style={thick},
  tick label style={/pgf/number format/fixed},
    every node near coord/.append style={font=\tiny}
]

 \addplot[lowyellow] [mark=triangle,mark size=2.7pt] coordinates {
 (1, 38.7) (2, 60.7) (3, 65.63)		
 };
 
\addplot[lowblue]  [mark=square]  coordinates {
(1, 32.2) (2, 52.43) (3, 58.94) 
};

\legend{ {E2H-base}, {T5-base}}
\end{axis}
\end{tikzpicture}
\end{minipage}%
}
\caption{Results of E2H-base and T5-base in low-resource scenarios. 
}
\label{fig:low}
\end{figure}
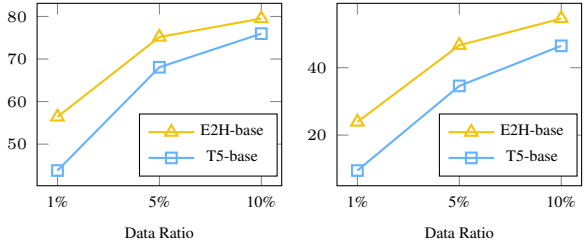
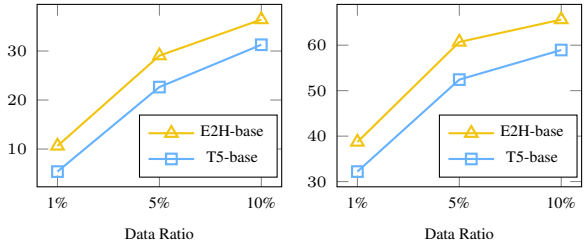

\subsection{Low-Resource Results}

Our experiments in low-resource scenarios show that E2H is particularly effective in situations where there is limited training data. As shown in  Figure \ref{fig:low}, by training on a fraction (1\%, 5\%, and 10\%) of the original data\footnote{ We repeat each experiment three times with different samples and report their averaged results.}, we observe that E2H-base significantly outperforms T5-base on all datasets. 
For example, when there is only 5\% of the training data, E2H-base obtains an average of 7.1, 12.0, 6.4, and 8.2 absolute points of improvement over T5-base on ACE04-Ent, ACE05-Rel, ACE05-E, and Rest14 respectively.  This highlights the effectiveness of our easy-to-hard learning framework when data is scarce. On one hand, the easy stage facilitates the model to identify the substructures of the target structure and capture the dependencies among them, which are difficult when there is limited data. On the other hand, the hard stage provides diverse and harder data to help the model tackle broad-range variations of the task, which is especially important in low-source scenarios. 

\begin{table}
\centering
\resizebox{0.49\textwidth}{!}
{
\setlength\tabcolsep{2pt}
\begin{tabular}{lccccc}
\toprule
\multirow{2}{*}{\textbf{Models}} & NER & RE & EE & ABSA \\
 &  \dataset{ACE04-Ent} & \dataset{ACE05-Rel} & \dataset{ACE05-E} & \dataset{Rest14}  \\
 \midrule
E2H-base & \bf{86.24} & \bf{65.44} & \bf{50.98} & \bf{75.40} \\
\quad w/o Skill$_1$ & 85.91 & 64.28 & 50.85 & 74.33 \\
\quad w/o Skill$_2$ & 86.13 & 64.05 & 49.89 & 74.98 \\
\quad w/o Skill$_3$ & - & 63.74 & - & 75.14 \\
\quad w/o Skill$_4$ & - & 64.00 & - & 74.88 \\
\bottomrule
\end{tabular}
}
\caption{Ablation results of E2H-base regarding different skills in the easy stage.}
\label{tab:ablation}
\vspace{-0.2cm}
\end{table}

\section{More Analysis}

\begin{table*}
\centering
\resizebox{0.9\textwidth}{!}
{
\begin{tabular}{lcccccc}
\toprule
\multirow{2}{*}{\textbf{Learning Strategy}} & \multirow{2}{*}{\textbf{Type}} & NER & RE & EE & ABSA & \multirow{2}{*}{Avg}\\
 &  & \dataset{ACE04-Ent} & \dataset{ACE05-Rel} & \dataset{ACE05-E} & \dataset{Rest14} &  \\
 \midrule
easy$\rightarrow$hard$\rightarrow$main & three-stage & \bf{86.24} & \bf{65.44} & \bf{50.98} & \bf{75.40} & \bf{69.52} \\
easy$\rightarrow$main$\rightarrow$hard & three-stage & 86.23 & 65.40 & 49.76 & 74.45 & 68.96 \\
easy+main+hard & multi-task & 86.10 & 64.46 & 49.16 & 73.94 & 68.42 \\
easy$\rightarrow$main & two-stage & 85.93 & 63.85 & 50.31 & 74.52 & 68.65 \\
hard$\rightarrow$main & two-stage & 85.99 & 64.41 & 49.26 & 74.67 & 68.58 \\
easy$\rightarrow$hard & two-stage & 86.18 & 65.35 & 46.69 & 75.34 & 68.39 \\
main & one-stage & 85.60 & 62.91 & 49.68 & 72.11 & 67.58 \\
\bottomrule
\end{tabular}
}
\caption{Experimental results of T5-base models trained with different learning strategies. The easy+main+hard strategy represents that the model is trained with the easy, main, and hard parts in a multi-task learning manner. The arrow $\rightarrow$ indicates the order between different stages.}
\label{tab:paradigm}
\end{table*}

\paragraph{Analysis on different learning strategies} 
In the main result table, we report the results of E2H trained with the easy$\rightarrow$hard$\rightarrow$main strategy, i.e., training the model in the easy, hard, and main stages sequentially. In this section, we investigate alternative learning strategies. Table \ref{tab:paradigm} reports the results of T5-base models trained with different learning strategies on four datasets across four tasks. We have the following observations: (1) The easy$\rightarrow$hard$\rightarrow$main strategy is the best among the seven concerned strategies. It performs better than other strategies on all datasets. (2) Easy-to-hard multi-stage learning outperforms multi-task learning (i.e., easy+main+hard). When the easy, main, and hard parts of the training data are used, the easy$\rightarrow$hard$\rightarrow$main and easy$\rightarrow$main$\rightarrow$hard strategies show superiority over the easy+main+hard strategy on all datasets. This indicates that easy-to-hard multi-stage learning is essential to the model's performance. (3) Each stage is critical to our E2H framework. Removing any of the stages will reduce the performance of E2H. (4) In general, three-stage learning is better than two-stage learning, and they are better than one-stage learning.

\paragraph{Is each skill necessary in the easy stage?}
To quantify the contribution of each skill, we examine the performance of E2H-base after removing a basic skill for training in the easy stage. Ablation results on four datasets across four tasks are shown in Table \ref{tab:ablation}. Removing any skill degrades the performance of E2H on the main task, indicating that recognizing substructures and the dependency between them is crucial to the model's performance.

\paragraph{Does easy-to-hard learning improve the model's cross-domain generalization ability?} 
To answer this question, we compare the performance of the E2H-base model and the T5-base model trained on a dataset on another dataset in a different domain of the same task. Table \ref{tab:domain} reports the results of the cross-domain generalization performance of different models on two dataset pairs: \dataset{CoNLL03}$\leftrightarrow$\dataset{ACE04-Ent} of the NER task and \dataset{Rest16}$\leftrightarrow$\dataset{Laptop14} of the ASTE task.
E2H-base performs better than T5-base in all scenarios. 
This indicates that easy-to-hard learning can enhance the model's cross-domain generalization ability. 

\begin{table}
\centering
\resizebox{0.49\textwidth}{!}
{
\setlength\tabcolsep{2pt}
\begin{tabular}{lccccc}
\toprule
\textbf{Models} &  \dataset{CoNLL03$\rightarrow$ACE04-Ent} & \dataset{ACE04-Ent$\rightarrow$CoNLL03 }   \\
\midrule
 T5-base & 19.54 & 17.45 \\
E2H-base & \bf{19.71} & \bf{30.08} \\
\midrule
\textbf{Models} &  \dataset{Rest16$\rightarrow$Laptop14} & \dataset{Laptop14$\rightarrow$Rest16}   \\
\midrule
T5-base & 42.37 & 60.50 \\
 E2H-base & \bf{44.86} & \bf{62.32} \\
\bottomrule
\end{tabular}
}
\caption{Cross-domain generalization performance of E2H-base and T5-base.}
\label{tab:domain}
\end{table}

\section{Related Work}

IE is a long-standing research area in natural language processing. Over the years, the paradigm for IE has undergone several transitions. Early approaches to IE focus on sequence labeling techniques \cite{mccallum-li-2003-early, ma-hovy-2016-end, ijcai2018p637, Li_Bing_Li_Lam_2019, zhang-etal-2021-cross}, in which each word in a text is assigned a label indicating its role in the extraction task. Span-based approaches \cite{luan-etal-2019-general, wang-etal-2020-pyramid, zhao-etal-2020-spanmlt, xu-etal-2021-learning, zhou-etal-2022-conner, zhou-etal-2023-improving}, which involve identifying spans in the text that correspond to the desired information, are later introduced for IE. MRC-based methods \cite{du-cardie-2020-event, li-etal-2020-unified, Mao_Shen_Yu_Cai_2021, xu-etal-2023-peerda} that frame the extraction task as a reading comprehension problem and generation-based methods \cite{BARTNER, lu-etal-2021-text2event, zhang-etal-2021-towards-generative} that generate the extracted information directly from the text have gained popularity in recent years for IE. They have been shown to be more effective and flexible. Most of these methods target a specific IE task. There have been some efforts to develop unified IE methods \cite{TANL, UIE, LasUIE}, which can unify various IE tasks with one framework. Our E2H framework, a unified IE framework, introduces a novel easy-to-hard learning paradigm for IE to reduce the gap between model and human learning.

From the perspective of improving the learning process, E2H shares similar spirits with transfer learning \cite{transfersurvey}, which uses the knowledge gained from solving one task to help solve another related task. By comparison, E2H learns basic skills specifically designed to assist with the target task. E2H is also related to curriculum learning \cite{curriculum, curriculumsurvey} in its fundamental motivation of learning from easy to hard. Curriculum learning, inspired by the human learning process, presents examples starting from the easiest samples, then gradually introducing more complex ones. However, curriculum learning involves the intricate task of ordering instances based on their difficulty. This requires a reliable difficulty criterion or a ranking system, which can be challenging to define and often necessitates substantial human effort. In contrast, E2H emphasizes on mastering certain fundamental skills prior to tackling more intricate tasks, eliminating the requirement for a difficulty criterion. This approach can be particularly beneficial in scenarios where the target task requires a distinct set of skills, or when the learning setting does not naturally provide a straightforward measure of difficulty.

\section{Conclusion}

This paper proposes an easy-to-hard learning framework consisting of the easy stage, the hard stage, and the main stage for IE. Two novel strategies are proposed to build the easy and hard parts of the framework to enable the learning process. Experimental results in both full and low-resource scenarios demonstrate the effectiveness of our framework and its superiority over one-stage learning methods.

\section*{Limitations}

While the results have shown the effectiveness of our framework in IE without using any additional resources,  we did not explore the potential enhancement by utilizing existing resources in the easy-to-hard learning process. On one hand, we can build the easy stage with the help of existing data of simpler tasks. 
On the other hand, the data of harder tasks can be used for the hard stage.
To enhance the E2H framework via effectively using existing resources is an interesting and promising direction. Another limitation is that we did not extensively explore the possible skill sets for each task. Exploring more approaches to obtain the skill sets is also open for future research. We plan to investigate these possibilities in our future work.

\bibliographystyle{acl_natbib}

\newpage
\appendix
\ADLinactivate 

\section{Appendix}
\subsection{Statistics of Datasets}
Statistics of datasets are reported in Table \ref{tab:details_datasets}. 
\label{sec:datasets}

\begin{table}[h]
  \centering
    \begin{tabular}{c|ccc}
    \toprule
           & \#Train & \#Val & \#Test \\
    \midrule
    CoNLL03    & 14,041  & 3,250  & 3,453  \\
    ACE04-Ent   & 6,202  & 745   & 812  \\
    ACE05-Ent    & 7,299  & 971   & 1,060  \\
    CoNLL04    & 922   & 231   & 288  \\
    ACE05-Rel   & 10,051  & 2,420  & 2,050  \\
    SciERC     & 1,861  & 275   & 551  \\
    ACE05-E   & 17,172 & 923 & 832  \\
    ACE05-E+   & 19,216  & 901   & 676  \\
    CASIE   & 11,189  & 1,778  & 3,208  \\
    Rest14    & 1,266  & 310   & 492  \\
    Laptop14  & 906   & 219   & 328  \\
    Rest15-ASTE      & 605   & 148   & 322  \\
    Rest16-ASTE     & 857   & 210   & 326  \\
    R-ACOS  & 1,530 & 171 & 583 \\
    L-ACOS  & 2,934 & 326 & 816 \\
    Rest15-ASQP & 834 & 209 & 537 \\
    Rest16-ASQP & 1,264 & 316 & 544 \\
    \bottomrule
    \end{tabular}%
  \caption{
    Statistics of datasets.
}
  \label{tab:details_datasets}
\end{table}

\subsection{Implementation Details}
\label{sec:implementation}
 We set the maximum input length to 384 and the maximum target length to 256. Following the practices of \citet{UIE}, we use a batch size of 64 for E2H-base and 32 for E2H-large. The learning rate is chosen from \{1e-4, 3e-4\} for E2H-base and \{5e-5, 1e-4\} for E2H-large, and we use the AdamW  optimizer \cite{AdamW} with linear learning rate decay. The number of training epochs for the easy, hard, and main stages are set to [15, 30, 30] or [25, 50, 50], with the easy stage having fewer epochs as it typically has more data. For the hard stage, we choose $M$ from \{1, 2\} for the datasets of the NER, RE, and EE tasks and from \{1, 2, 3\} for the datasets of the ABSA task. The parameters are chosen based on the model's performance on the development set. Generally, for large datasets such as ACE05-E, a smaller value of $M$ like 1 is more appropriate, while for smaller datasets such as Laptop14, a larger value of $M$ such as 3 is preferred. All experiments are conducted on NVIDIA Tesla A100.

\subsection{Examples of IE tasks}
\label{sec:examples}
Detailed examples of different IE tasks are shown in Tables \ref{tab:ner_examples}-\ref{tab:asqp_examples}. We use the structural extraction language proposed by \citet{UIE} to encode the target structure. 

\begin{table*}[htb]
\centering
\resizebox{\textwidth}{!}
{
\begin{tabular}{c| p{0.54\linewidth}| p{0.42\linewidth}}
\toprule
\textbf{Task} & \textbf{Input} & \textbf{Target} \\
\midrule
NER & \red{[HEC] [HES]} \blue{[Ent] location [Ent] miscellaneous [Ent] organization [Ent] person} [Text] Only France and Britain backed Fischler's proposal. & \sel{((location:~France) (location:~Britain) (person:~Fischler))} \\
\midrule
Skill$_1$  & \red{[HEC]} \blue{[Ent] location [Ent] miscellaneous [Ent] organization [Ent] person} [Text] Only France and Britain backed Fischler's proposal. & \sel{((location) (person))}  \\
\midrule
Skill$_2$ & \red{[HEC] [HES]} \brown{[Ent] location} [Text] Only France and Britain backed Fischler's proposal. & \sel{((location:~France) (location:~Britain))} \\
\bottomrule
\end{tabular}
}
\caption{Detailed Examples for NER. We provide an instance for the main task and each skill. We highlight \hint{Hint} in \red{red}, \hint{Constraint} in \brown{brown}, and \hint{Schema} in \blue{blue}. \red{[HEC]} and \red{[HES]} are the entity category hint and entity span hint, respectively. [Ent] is a special token to denote the entity category.}
\label{tab:ner_examples}
\end{table*}

\begin{table*}
\centering
\resizebox{\textwidth}{!}
{
\begin{tabular}{c| p{0.54\linewidth}| p{0.42\linewidth}}
\toprule
\textbf{Task} & \textbf{Input} & \textbf{Target} \\
\midrule
RE & \red{[HE] [HR]} \blue{[Ent] generic [Ent] material [Ent] method [Ent] metric [Ent] other scientific term [Ent] task [Rel] compare [Rel] conjunction [Rel] evaluate for [Rel] feature of [Rel] hyponym of [Rel] part of [Rel] used for} [Text] The demonstrator embodies an interesting combination of hand-built, symbolic resources and stochastic processes. & \sel{((task:~demonstrator) (material:~hand-built, symbolic resources (part of:~demonstrator)(conjunction:~stochastic processes)) (method:~stochastic processes (part of:~demonstrator)))} \\
\midrule
Skill$_1$ & \red{[HE]} \blue{[Ent] generic [Ent] material [Ent] method [Ent] metric [Ent] other scientific term [Ent] task} [Text] The demonstrator embodies an interesting combination of hand-built, symbolic resources and stochastic processes. & \sel{((task:~demonstrator) (material:~hand-built, symbolic resources) (method:~stochastic processes))} \\
\midrule
Skill$_2$ & \red{[HE] [HR]} \brown{[Ent] method:~stochastic processes} \blue{[Rel] compare [Rel] conjunction [Rel] evaluate for [Rel] feature of [Rel] hyponym of [Rel] part of [Rel] used for} [Text] The demonstrator embodies an interesting combination of hand-built, symbolic resources and stochastic processes. & \sel{((method:~stochastic processes (part of:~demonstrator)))}
 \\
 \midrule
 Skill$_3$ & \red{[HR]} \blue{[Rel] compare [Rel] conjunction [Rel] evaluate for [Rel] feature of [Rel] hyponym of [Rel] part of [Rel] used for} [Text] The demonstrator embodies an interesting combination of hand-built, symbolic resources and stochastic processes. & \sel{((part of) (conjunction))} \\
 \midrule
Skill$_4$ & \red{[HE] [HR]} \brown{[Rel] conjunction} \blue{[Ent] generic [Ent] material [Ent] method [Ent] metric [Ent] other scientific term [Ent] task} [Text] The demonstrator embodies an interesting combination of hand-built, symbolic resources and stochastic processes. & \sel{((material:~hand-built, symbolic resources (conjunction:~stochastic processes)))} \\
 \bottomrule
\end{tabular}
}
\caption{Detailed Examples for RE. We provide an instance for the main task and each skill. We highlight \hint{Hint} in \red{red}, \hint{Constraint} in \brown{brown}, and \hint{Schema} in \blue{blue}. \red{[HE]} and \red{[HR]} are the entity hint and relation hint, respectively. [Ent] and [Rel] are special tokens to denote the entity category and relation, respectively.}

\label{tab:re_examples}
\end{table*}

\begin{table*}[htb]
\centering
\resizebox{\textwidth}{!}
{
\begin{tabular}{c| p{0.7\linewidth}| p{0.26\linewidth}}
\toprule
\textbf{Task} & \textbf{Input} & \textbf{Target} \\
\midrule
EE & \red{[HT] [HA]} \blue{[Tri] acquit [Tri] appeal [Tri] arrest jail [Tri] attack [Tri] born [Tri] charge indict [Tri] convict [Tri] declare bankruptcy [Tri] demonstrate [Tri] die [Tri] divorce [Tri] elect [Tri] end organization [Tri] end position [Tri] execute [Tri] extradite [Tri] fine [Tri] injure [Tri] marry [Tri] meet [Tri] merge organization [Tri] nominate [Tri] pardon [Tri] phone write [Tri] release parole [Tri] sentence [Tri] start organization [Tri] start position [Tri] sue [Tri] transfer money [Tri] transfer ownership [Tri] transport [Tri] trial hearing [Arg] adjudicator [Arg] agent [Arg] artifact [Arg] attacker [Arg] beneficiary [Arg] buyer [Arg] defendant [Arg] destination [Arg] entity [Arg] giver [Arg] instrument [Arg] organization [Arg] origin [Arg] person [Arg] place [Arg] plaintiff [Arg] prosecutor [Arg] recipient [Arg] seller [Arg] target [Arg] vehicle [Arg] victim} [Text] It was talking something about the war in Iraq. I guess it's a good thing about the elections that are going on.  & \sel{((attack:~war (place:~Iraq)) (elect:~elections (place:~Iraq)))} \\
\midrule
Skill$_1$  & \red{[HT]} \blue{[Tri] acquit [Tri] appeal [Tri] arrest jail [Tri] attack [Tri] born [Tri] charge indict [Tri] convict [Tri] declare bankruptcy [Tri] demonstrate [Tri] die [Tri] divorce [Tri] elect [Tri] end organization [Tri] end position [Tri] execute [Tri] extradite [Tri] fine [Tri] injure [Tri] marry [Tri] meet [Tri] merge organization [Tri] nominate [Tri] pardon [Tri] phone write [Tri] release parole [Tri] sentence [Tri] start organization [Tri] start position [Tri] sue [Tri] transfer money [Tri] transfer ownership [Tri] transport [Tri] trial hearing} [Text] It was talking something about the war in Iraq. I guess it's a good thing about the elections that are going on. & \sel{((attack:~war) (elect:~elections))}  \\
\midrule
Skill$_2$ & \red{[HT] [HA]} \brown{[Tri] attack:~war} \blue{[Arg] adjudicator [Arg] agent [Arg] artifact [Arg] attacker [Arg] beneficiary [Arg] buyer [Arg] defendant [Arg] destination [Arg] entity [Arg] giver [Arg] instrument [Arg] organization [Arg] origin [Arg] person [Arg] place [Arg] plaintiff [Arg] prosecutor [Arg] recipient [Arg] seller [Arg] target [Arg] vehicle [Arg] victim}  [Text] It was talking something about the war in Iraq. I guess it's a good thing about the elections that are going on. & \sel{((attack:~war (place:~Iraq)))} \\
\bottomrule
\end{tabular}
}
\caption{Detailed Examples for EE. We provide an instance for the main task and each skill. We highlight \hint{Hint} in \red{red}, \hint{Constraint} in \brown{brown}, and \hint{Schema} in \blue{blue}. \red{[HT]} and \red{[HA]} are the event trigger hint and event argument hint, respectively. [Tri] and [Arg] are special tokens to denote the event category and argument category, respectively.}
\label{tab:ee_examples}
\end{table*}

\begin{table*}
\centering
\resizebox{\textwidth}{!}
{
\begin{tabular}{c| p{0.47\linewidth}| p{0.46\linewidth}}
\toprule
\textbf{Task} & \textbf{Input} & \textbf{Target} \\
\midrule
ASTE & \red{[HE] [HR]} \blue{[Ent] aspect [Ent] opinion [Rel] negative [Rel] neutral [Rel] positive} [Text] Great food but the service was dreadful! & \sel{((opinion:~Great) (aspect:~food (positive:~Great)) (aspect:~service (negative:~dreadful)) (opinion:~dreadful))} \\
\midrule
Skill$_1$ & \red{[HE]} \blue{[Ent] aspect [Ent] opinion} [Text] Great food but the service was dreadful! & \sel{((opinion:~Great) (aspect:~food) (aspect:~service) (opinion:~dreadful))} \\
\midrule
Skill$_2$ & \red{[HE] [HR]} \brown{[Ent] aspect:~sevice} \blue{[Rel] negative [Rel] neutral [Rel] positive} [Text] Great food but the service was dreadful! & \sel{((aspect:~service (negative:~dreadful)))} 
 \\
 \midrule
 Skill$_3$ & \red{[HR]} \blue{[Rel] negative [Rel] neutral [Rel] positive} [Text] Great food but the service was dreadful! & \sel{((positive) (negative))} \\
 \midrule
Skill$_4$ & \red{[HE] [HR]} \brown{[Rel] positive} \blue{[Ent] aspect [Ent] opinion} [Text] Great food but the service was dreadful! & \sel{((aspect:~food (positive:~Great)))} \\
 \bottomrule
\end{tabular}
}
\caption{Detailed Examples for ASTE. We provide an instance for the main task and each skill. We highlight \hint{Hint} in \red{red}, \hint{Constraint} in \brown{brown}, and \hint{Schema} in \blue{blue}. 
Following \citet{UIE}, we formulate ASTE as the RE task, where aspect terms and opinion terms are entities, and sentiment polarities are relations.
\red{[HE]} and \red{[HR]} are the entity hint and relation hint, respectively. [Ent] and [Rel] are special tokens to denote the entity category and relation, respectively.}

\label{tab:aste_examples}
\end{table*}

\begin{table*}
\centering
\resizebox{\textwidth}{!}
{
\begin{tabular}{c| p{0.47\linewidth}| p{0.46\linewidth}}
\toprule
\textbf{Task} & \textbf{Input} & \textbf{Target} \\
\midrule
ASQP & \red{[HC] [HA]} \blue{[Cat] category [Arg] aspect [Arg] opinion [Arg] polarity} [Text] The pizza is delicious. & \sel{((category:~food quality (aspect:~pizza) (opinion:~delicious) (polarity:~positive))} \\
\midrule
Skill$_1$ & \red{[HC]} \blue{[Cat] category} [Text] The pizza is delicious. & \sel{((category:~food quality))} \\
\midrule
Skill$_2$ & \red{[HC] [HA]} \blue{[Cat] category [Arg] aspect} [Text] The pizza is delicious. & \sel{((category:~food quality (aspect:~pizza))} \\
 \midrule
 Skill$_3$ & \red{[HC] [HA]} \blue{[Cat] category [Arg] opinion} [Text] The pizza is delicious. & \sel{((category:~food quality (opinion:~delicious))} \\
 \midrule
Skill$_4$ & \red{[HC] [HA]} \blue{[Cat] category [Arg] polarity} [Text] The pizza is delicious. & \sel{((category:~food quality (polarity:~positive))} \\
 \bottomrule
\end{tabular}
}
\caption{Detailed Examples for ASQP. We provide an instance for the main task and each skill. We highlight \hint{Hint} in \red{red}, \hint{Constraint} in \brown{brown}, and \hint{Schema} in \blue{blue}. 
We treat the aspect term, opinion term, and sentiment polarity as the arguments of the aspect category.
\red{[HC]} and \red{[HA]} are the aspect category hint and argument hint, respectively. [Cat] and [Arg] are special tokens to denote the aspect category and its arguments, respectively.}

\label{tab:asqp_examples}
\end{table*}

\end{document}